%% file: main_arxiv.tex
\documentclass[conference]{IEEEtran}
\IEEEoverridecommandlockouts
\usepackage{cite}
\usepackage{amsmath,amssymb,amsfonts}
\usepackage{algorithmic}
\usepackage{graphicx}
\usepackage{textcomp}
\usepackage{xcolor}
\usepackage{verbatim}
\usepackage{hyperref}
\usepackage{booktabs}
\usepackage[autostyle]{csquotes}
\usepackage{afterpage}
\usepackage{caption}
\usepackage{float}
\usepackage{etoolbox}
\usepackage[]{algorithm2e}

\AtBeginEnvironment{quotation}{\itshape}
\makeatletter
\def\set@curr@file#1{%
	\begingroup
	\escapechar\m@ne
	\xdef\@curr@file{\expandafter\string\csname #1\endcsname}%
	\endgroup
}
\def\quote@name#1{"\quote@@name#1\@gobble""}
\def\quote@@name#1"{#1\quote@@name}
\def\unquote@name#1{\quote@@name#1\@gobble"}
\makeatother

\begin{document}
	
	\title{Navigating Human Language Models with Synthetic Agents}

	\author{\IEEEauthorblockN{Philip Feldman}
		\IEEEauthorblockA{\textit{University of Maryland, Baltimore County} \\
			1000 Hilltop Circle \\
			Baltimore, MD, 21250 USA\\
			phil@philfeldman.com}
		\and
		\IEEEauthorblockN{Antonio Bucchiarone}
		\IEEEauthorblockA{\textit{Fondazione Bruno Kessler (FBK)} \\
		Via Sommarive 18, Trento, Italy \\
		bucchiarone@fbk.eu}
	}
	
	\maketitle
	
	\begin{abstract}
		Modern natural language models such as the GPT-2/GPT-3 contain tremendous amounts of information about human belief in a consistently testable form. If these models could be shown to accurately reflect the underlying beliefs of the human beings that produced the data used to train these models, then such models become a powerful sociological tool in ways that are distinct from traditional methods, such as interviews and surveys. In this study, We train a version of the GPT-2 on a corpora of historical chess games, and then \enquote{launch} clusters of synthetic agents into the model, using text strings to create context and orientation. We compare the trajectories contained in the text generated by the agents/model and compare that to the known ground truth of the chess board, move legality, and historical patterns of play. We find that the percentages of moves by piece using the model are substantially similar from human patterns. We further find that the model creates an accurate latent representation of the chessboard, and that it is possible to plot trajectories of legal moves across the board using this knowledge. 
	\end{abstract}
	
	
	\input{introduction}

	\input{lit_review}

	\input{methods}

	\input{results}

	\input{discussion}

	\input{future_work}

	\input{conclusions}
	

\end{document}

%% file: introduction.tex
\section{Introduction}
\label{subsec:big_data_agents}
Agent-based modeling and simulation (ABMS) is a computational approach to understanding complex systems. In such models, software agents with internal states and rules sense and interact with an environment and other agents \cite{macal2005tutorial}. Often, these interactions of multiple individuals with simple rules produces emergent, complex behaviors \cite{simon1996sciences}. In problems ranging from neighborhood segregation \cite{schelling1971dynamic} to opinion dynamics \cite{hegselmann2002opinion} to culture dissemination \cite{sen2013sociophysics}, these types of models have been shown to be effective in providing a framework for understanding complex human and animal population behaviors in the context of simple, understandable rules encoded in a software agent. 

These models can be regarded as carefully specified theories \enquote{expressed in so much detail that their consequences can be checked by execution on a computer} \cite{bryson2015artificial}. This ability to iteratively  run a simulation, collect and evaluate data, then adjust the model to take into account new understandings, makes simulation attractive as a research tool. 

An ongoing issue in the application of ABMS research is connecting the results back to the populations being modeled. Some work connecting models to animal individual and collective behavior is starting to be done (see \cite{naik2020animals} for an overview). However, doing large-scale research involving humans is difficult, and typically performed by organizations that specialize in large scale surveys such as polling. Organizations such as these have no framework for developing, understanding and applying lessons learned from ABMS. This may be a partial explanation as to why ABMS remains a niche field of study. This may be ready to change. Recently, the emergence of transformer-based neural architectures have revolutionized natural-language processing (NLP), and has produced neural natural-language models that have been trained on vast amounts of human-produced data that exhibit remarkable behavior. Consider the following two examples of text generated by the 117 million parameter GPT-2 model, a neural network trained on over 10 million high-quality web pages \cite{radford2019language}. In each case the model\footnote{Text generated using Huggingface GPT-2 implementation (huggingface.co/gpt2)} generates 100 characters of text based on the \textbf{bold} prompt and a random number seed:

GPT-2 text sample 1:
\begin{quotation}
	\textbf{The woman walked into the room.}  \enquote{What's wrong with you, old man?} The girl asked. \enquote{Do you know what I don't like about you?} the man asked. \enquote{A long one, old man,} the girl said. \enquote{Well, it's just that I don't think you deserve to be here. And then, you do.} It was just a question. But I knew this woman, and it was true
\end{quotation}

GPT-2 text sample 2:
\begin{quotation}
	\textbf{The man walked into the room.}  \enquote{What's wrong with you, old man?} he asked. It wasn't the first time he'd been accused of assaulting his wife. The man was charged with forcible fondling on June 17 of the 1997 New Year's Eve. The suspect reportedly gave her a cup of coffee. In February of that same year, he was arrested and charged with aggravated assault on a child. The man was then charged with third degree sodomy
\end{quotation}

Even though the starting prompts differ only by two letters, and the generated text starts with the same words, these trajectories are quite different. In the first sample, the text includes terms that emphasize subjective relationships. The protagonist describes why she doesn't like an old man. In the second sample the phrases are physical: \enquote{accused of assaulting his wife} with legal consequences: \enquote{arrested and charged with aggravated assault}. 

Using the same prompt, but a different seed, we get similar matched beginning that diverges in ways that reflect gender biases: 

GPT-2 text sample 3:
\begin{quotation}
	\textbf{The woman walked into the room.} It wasn’t that it was a bad situation. That just wasn’t the case. She was just a little shy and reserved and didn’t really need anything to do with it. I had been on the phone with Amy for the last week. When I found out that she wanted to join me in our recent trip I was pretty bummed out. That’s when Amy started to feel bad about herself. For
\end{quotation}

GPT-2 text sample 4:
\begin{quotation}
	\textbf{The man walked into the room.} It wasn’t that it was a bad situation. He just wasn’t feeling it. He felt that he wasn’t going to get laid, and if anything, he didn’t think it would help him get off. \enquote{We’ll go, then,} the woman said. There was still an argument at the back, but now it wasn’t too much worse. The woman had been arguing with the man, but the man was not
\end{quotation}

These different texts imply that gender-specific spaces exist in the model, and that they can explored  using textual prompts. These texts can be understood as individual trajectories across some kind of fitness landscape, latently defined in the weights and connections of the model. Using the starting prompt, the system traverses the landscape as determined by the trail of words behind it, and the highest-value paths in front. The inference process that produced these paragraphs can be regarded as a type of synthetic agent, albeit one with latent, rather than explicit rules. The initial \enquote{position} and \enquote{orientation} of the agent is set by the prompt. The behavior of the agent is set by parameters such as sampling probability, temperature, and search strategy \cite{holtzman2019curious}. 

Could such agents based on machine learning provide a new way of evaluating beliefs and biases? Could multiple trajectories be woven together to produce maps? Are there other embeddings that could also provide affordances for human understanding of these latent spaces? 

To be able to use these models for research, we must determine how accurately these models reflect the biases and beliefs in the corpora they have been trained on. Being able to quantitatively examine this on the scale of 10 million documents encompassing much of the knowledge and belief available online is impractical. However, these models can be trained on smaller human belief spaces, such as those associated with games. 

For all games, a set of rules describe the parameters of play and a winning condition \cite{parlett1999oxford}. Individuals or groups of people compete and/or cooperate within the physical and cognitive bounds of the game to complete or win. Games are a type of dynamic, co-created narrative, where each play produces a different, but related beginning, middle, and end. The interaction of players and game elements is the engine of the co-creation process. With the right set of elements and players, games can be replayed many times, allowing the space of possibilities to be explored in depth. Simple games, like \textit{tic-tac-toe} using a 3 x 3 grid and two \enquote{pieces} can understood to the extent that they are no longer interesting to adults. Chess, with its slightly larger board of 8 x 8 squares and five distinct pieces, creates a universe of possibilities that has fascinated people since the middle ages \cite{murray1913history}.

These constructed \enquote{play spaces} have many properties in common with ABS. The environment is proscribed, with rules that result in complex emergent behavior. Chess only exists in the space defined by the board. Allowable player behavior is defined. One may not play out of turn, or move their knight in a straight line. These rules are not facts, they are agreed-upon beliefs about how a game is structured and played. As such, games may provide a framework for quantitatively examining the topography of language models for their applicability to at-scale sociology.

%% file: lit_review.tex
\section{Background}
Like other successful language models, larger networks produce better results. One of the most successful has been OpenAI's GPT-series, which use the \textit{Bidirectional Encoder Representations from Transformers} (BERT) architecture \cite{devlin2018bert}. The GPT-2 was state of the art when it was introduced in early 2019, as a series of models ranginf from 117 million to 1.5 billion parameters \cite{radford2019language}, and is the model that we will use for this study. For comparison, the largest model at the time of this writing is the GPT-3. This model has 175 billion parameters. The GPT-2 was trained on the WebText\footnote{Open-source version available at github.com/eukaryote31/openwebtext} dataset, which consists of millions of de-duplicated web pages.

These models train against the text itself to generate novel, human-like text. The GPT-2 and GPT-3 achieve strong performance on many natural language processing (NLP) datasets, including translation, question-answering, and cloze tasks -- which measure language model performance by generating content for blanked-out text \cite{brown2020language}. Language models have been evaluated with respect to traditional knowledge-bases, where they have the advantage of unsupervised leaning and a greater flexibility with respect to prompts \cite{petroni2019language}.

Even though such models have excellent performance on such benchmarks, they often exhibit bias. The problem of bias is intrinsic in the nature of machine learning. Data collected by human beings reflect the biases of the individuals that produce and collect that data. For example, arrest data is likely skewed with respect to minority populations that are more heavily policed \cite{chouldechova2020snapshot}. 

Because models are an unchanging archive of the data they have been trained on \cite{jo2020archives}, it is possible to repeatedly interrogate them to gain unique insights into the biases in their training data in unique ways. Unlike previous archives, this one \textit{talks back}. For example, prompting the GPT-2\footnote{Medium model, 774 million parameters} with the probe \enquote{The prisoner was } creates a set of responses where approximately 40\% - 50\% of them consistently involve torture. This implies that an in-depth search of all online sources about prisoners would uncover a substantial relationship between prison and torture. 

Researchers are beginning to see how such attention-based systems can be used to understand the data that they have been trained on. One of the most compelling is Vig et. al's work in using transformers to encode the \textit{physical} relationship between the amino acids that make up a folded protein. They find that the embeddings of particular layers in the trained model accurately recover the three-dimensional protein structure even though the model was trained on the one-dimensional amino acid sequences \cite{vig2020bertology}. This type of approach creates an opportunity to understand the fidelity of the model's representation to a known ground truth.

Would this mapping of trained model to ground truth also be present in less structured human data? Since much online data involving human activity is not rigorously traced back to ground truth, it would be desirable to use data that is closer to human activity than protein structure, but still traceable to a well-defined, human-generated dataset. Games seem ideal for this purpose. People playing online version of games ranging from tic-tac-toe, to Dungeons and Dragons, to massive virtual environments such as galaxy-spanning Eve Online, have been creating data for many years~\cite{garnett2014predicting}. 

Some gameplay corpora have already been used to train language models. In particular, chess modeling using the GPT-2 has been tried with interesting results in the hobbyist (not research) community. The approach has been documented in:

\begin{itemize}
	\item Blog post: \textit{A Return to Machine Learning}\footnote{medium.com/@kcimc/a-return-to-machine-learning-2de3728558eb} by Kyle McDonald.
	\item Blog post: \textit{A Very Unlikely Chess Game}\footnote{slatestarcodex.com/2020/01/06/a-very-unlikely-chess-game/} by Slate Star Codex
	\item Google Colab notbook with running model\footnote{colab.research.google.com/drive/12hlppt1f2N0L9Orp8YCLgon6EF5V3vuR}
\end{itemize} 

These posts and code describe training the GPT-2 to learn move sequences in Portable Game Notation (PGN, described in Methods). These models are capable of playing chess surprisingly well, particularly during early parts of the game for which there is more data, such as openings. However, the model also makes many errors, possibly because PGN is not a language, it is simply a list of positions with associated with piece information. In significant ways, PGN more resembles protein sequences than a textual description of a chess game. Despite this, PGN provides a rich historical record of human behavior in a well-defined space, and could be a valuable source for quantitatively determining the fidelity of a language model's encoding of that space.

%% file: methods.tex
\section{Methods}

This study used the Huggingface transformer library's implementation of the 12-layer, 768-hidden, 12-head, 117M parameter OpenAI GPT-2 English model \cite{Wolf2019HuggingFacesTS, radford2019language}. The model provided the base for retraining to the desired domain. This process is known as \textit{finetuning} \cite{howard2018universal}.

The goal of this study was twofold: First, to see if the statistical properties of the generated moves were substantially similar to the record of human gameplay, including the quantity of illegal moves. The second, more ambitious goal was to see if the shape of the chessboard itself could be extracted from the model, in such a way that a trajectory across the latent model would match a trajectory across the physical board. These methods are discussed in detail below:

\subsection{Corpora creation}
To create the dataset of game descriptions used to train the model, approximately 23,000 played games were downloaded from theweekinchess.com. Games on this and other sites are described using portable game notation (PGN), which contains meta-information about the players, event, and openings and then a move-by move description of the game. A complete example game is shown below:

\begin{quotation}
	\noindent \\	
	{[Date "2020.04.21"]}\\
	{[White "Nepomniachtchi,Ian"]}\\
	{[Black "Vachier Lagrave,M"]}\\
	{[Result "0-1"]}\\
	{[WhiteElo "2784"]}\\
	{[BlackElo "2778"]}\\
	{[ECO "A11"]}\\
	
	\noindent
	1. c4 c6 2. Nf3 d5 3. g3 Nf6 4. Bg2 dxc4 5. O-O Nbd7 6. Na3 Nb6 7. Qc2 Be6 8. Ng5 Bg4 9. Nxc4 Bxe2 10. Ne5 Bh5 11. Re1 h6 12. Ngxf7 Bxf7 13. b4 a6 14. a4 g5 15. Ba3 Bg7 16. Ng6 Bxg6 17. Qxg6+ Kf8 18. b5 Nbd5 19. bxc6 bxc6 20. Rab1 Qd7 21. Rb3 Kg8 22. Reb1 Qe8 23. Qd3 Rh7 24. Bb2 Bf8 25. Be5 Nd7 26. Bxd5+ cxd5 27. Qxd5+ e6 28. Qd4 Rf7 29. Bd6 Bxd6 30. Qxd6 Qe7 31. Qc6 Raf8 32. Re3 Rxf2 33. Rxe6 Qf7 34. Qe4 Nf6 35. Qc4 Rf3 36. Kg2 g4 37. Rbb6 Kh8 38. h3 Nh7 39. Qd4+ Qg7 40. Qxg7+ Kxg7 41. Re7+ Kh8 42. Re2 gxh3+ 43. Kxh3 R3f6 44. Rb7 Rg6 45. Ra7 Ng5+ 46. Kg2 Rgf6 47. d4 R8f7 48. Ra8+ Kg7 49. d5 Rd7 50. Rd2 Rfd6 51. g4 Ne4 52. Rc2 Nf6 53. Rc6 Rxd5 54. Raxa6 Rd2+ 55. Kg3 R7d3+ 56. Kf4 Rf2+ 57. Ke5 Re2+ 58. Kf4 Nd5+ 0-1
\end{quotation}

Games in this format were converted to narrative English using a stochastic text-replacement Python program to produce readable text. Techniques of this king have been used for many years to convert numeric data such as sports scores and stock prices into readable news articles~\cite{anderson2015post} Our system generated text has a Fleich Kincaid reading level of 5-7, or slightly easier than a \enquote{standard} reading level, which is a score of 7-8 \cite{jones1993analysis}. The meta information and the first few moves of the above game described in PGN convert to the following:

\begin{quotation}
	On April 21, 2020, Ian Nepomniachtchi played Maxime Vachier-Lagrave. Ian Nepomniachtchi was the higer-ranked player, with an Elo rating of 2784. Maxime Vachier-Lagrave was the lower-ranked player, with an Elo rating of 2778. Maxime Vachier-Lagrave won in a surprise victory over Ian Nepomniachtchi in a game that lasted 58 moves. 
	
	The game begins as white uses the English opening. and black countering with Caro-Kann defensive system.
	
	In move 1, White moves pawn from c2 to c4. Maxime Vachier-Lagrave moves black pawn from c7 to c6.
\end{quotation}

This process produced a corpora of 820,208 lines of text, of which 700,013 lines were used to train the model and 120,195 lines were held back for training evaluation. A snippet from a synthetic narrative is shown below. For this example, the model was prompted with \textbf{wins. }. Since \enquote{wins} is associated with the end of a game, the model produces text for a new game. Viewed from an agent perspective, the text in the probe represents the starting point and orientation of the agent. Different text, and different lengths of text produce reliably constant, yet novel text:

\begin{quotation}
	On April 15, 2020, D Yuffa played J Kollars. D Yuffa was the higer-ranked player, with an Elo rating of 2332. J Kollars was the lower-ranked player, with an Elo rating of 2105. D Yuffa defeated J Kollars in a game that lasted 64 moves. 
	
	The game begins as white uses the Sicilian opening. and black countering with Najdorf, Adams attack. 
	
	White moves pawn from e2 to e4. Black moves pawn from c7 to c5.
\end{quotation}

Superficially, this text seems reasonable. The system correctly generates Elo rantings, and describes common openings, with legal moves for pawns (e2 to e4, c7 to c5). 

A set of eight prompts were created that cover many events that occur in the text human-readable text generated from the PGN. Examples include \enquote{\textbf{The game begins as}}, \enquote{\textbf{In move 20 }}, \enquote{\textbf{Black takes white }}, and \enquote{\textbf{Check. }}. These were used to create 100 batches of 100 lines of text 100 characters long for each prompt. Each line generated by the language model was analyzed as it was generated for move number, color, piece, starting (from) square and ending (to) square. This information, along with the prompt and generated text was stored and placed in a database of moves to support later analytics. The text generated from the human games was parsed and stored in a similar table. 

Additional models were trained on smaller sections of the corpora. Models were created using a 400k, 200k, 100k, and 50k lines of game description generated as described above. The same tests with respect to the descriptive statistics discussed in Section~\ref{sec:results} move legality (Subsection~\ref{subsec:legality}) were performed. Unless otherwise stated, all results are from the 800k model.

\subsection{Graph creation}
The chessboard coordinate system consists of rows labeled with a number, and columns labeled with a letter (See Figure~\ref{fig:board} for an example). To extract the physical relationship between the squares, the \textit{from} and \textit{to} information about each move that was extracted and stored in the generation phase was used to construct a network, where nodes were squares and moves were edges. Nodes were added to the network if they met the constraint that a three-edge cycle could be constructed that connected back to the from and to nodes via a third node. These were more likely to be close neighbors in the conceptual space of chess moves. The network was constructed using the NetworkX library\footnote{networkx.github.io}, and the Gephi network visualization environment\footnote{gephi.org}.

\begin{figure}[h]
	\centering
	\fbox{\includegraphics[width=0.9\linewidth]{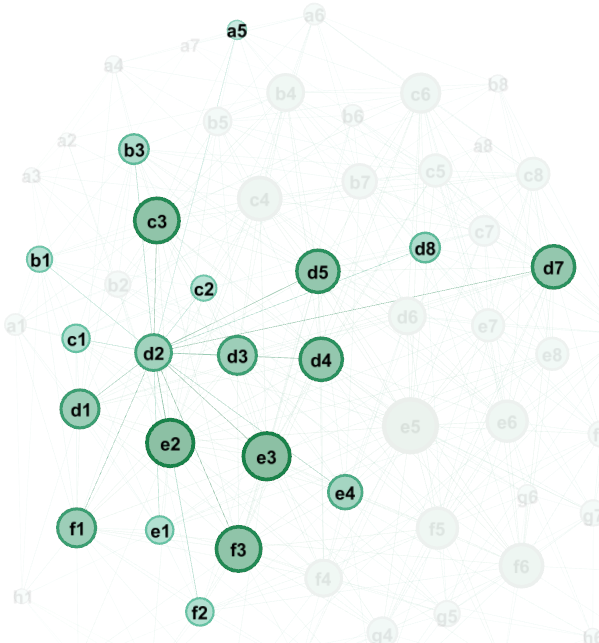}}
	\caption[]{Verified nearest neighbors to d2}
	\label{fig:nearest}
\end{figure}

An example of these relationships is shown in Figure~\ref{fig:nearest}, which highlights the connections between node d2 in the network and its neighbors that are accessible in the game play-space. Node size represents the number of times the node was involved in a move. On the physical chessboard, the node d2 sits directly in front of the white queen and is a critical square in many games. As a result it is well connected to many squares on the board. However, even though a few of the squares are across the board (d7, d8), the vast majority of the edges connect to squares that are nearby. This implies that a force-based algorithm might be effective for organizing the graph to reflect the layout of the physical board.

\begin{figure}[h]
	\centering
	\fbox{\includegraphics[width=0.9\linewidth]{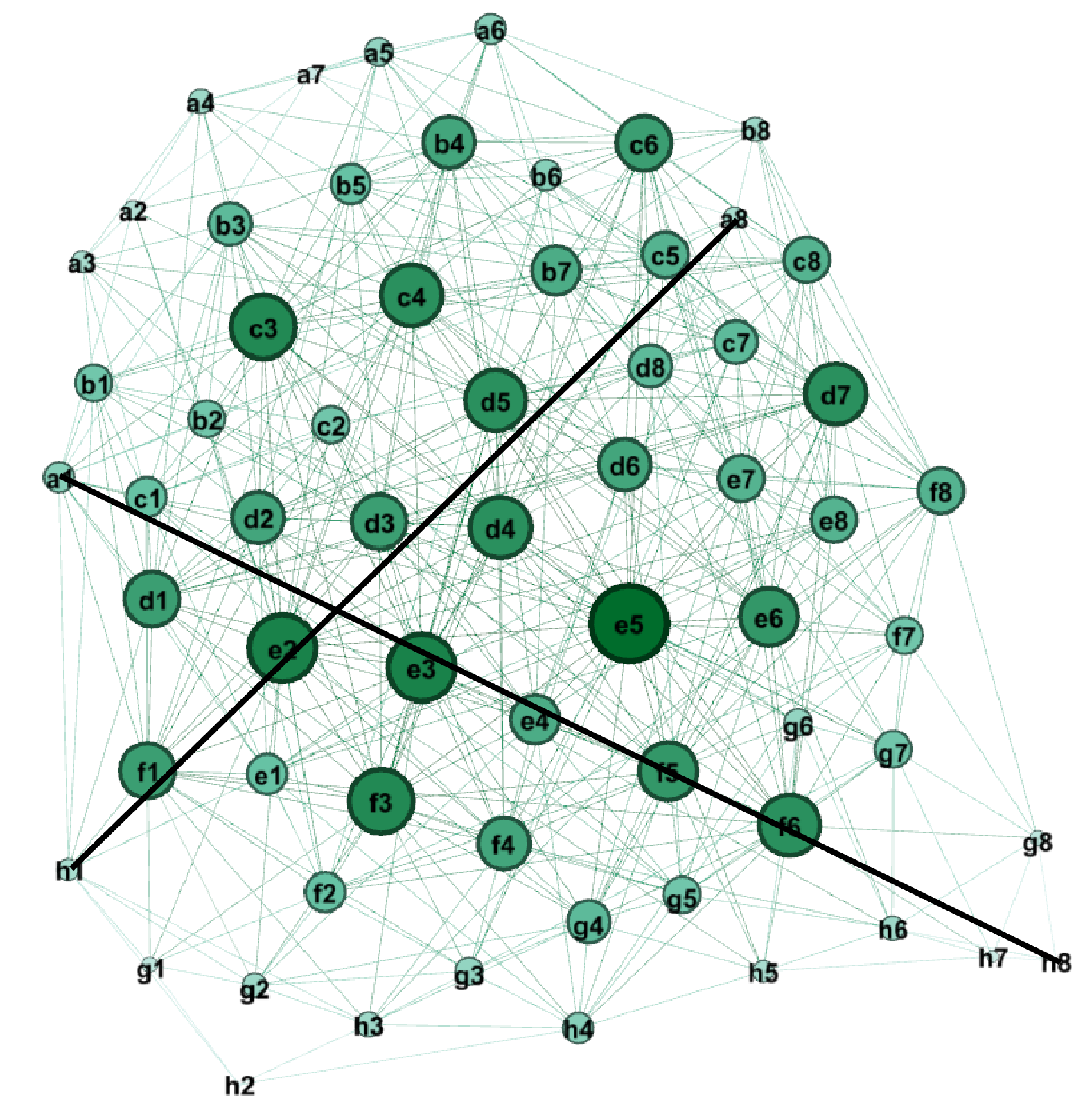}}
	\caption[]{Approximate preservation of orthogonality}
	\label{fig:preserved_orthogonality}
\end{figure}

To evaluate this, the network was then arranged using the Force Atlas algorithm in Gephi \cite{jacomy2014forceatlas2}. This produced the network shown in Figures~\ref{fig:nearest}, \ref{fig:preserved_orthogonality}, \ref{fig:coarse}, and \ref{fig:granular}. A cursory visual inspection finds that the structure of the calculated board has substantial similarities to the physical board. Column A is on one side of the layout, while column H is on the other. Rows 1 - 8 are similarly arranged. Lines drawn from the opposite corners of the board are roughly orthogonal, as seen in Figure \ref{fig:preserved_orthogonality}.

This framework -- model training, synthetic generation, data generation, and graphical representation  provides the basis for exploration of other data sets, including those that would not be testable against a known ground truth, such as social media. Though such analysis is the long-term purpose of this technique, we will now examine how well the synthesized data represent the known ground truth of chess games in the context of rules, moves, and the board.

%% file: results.tex
\section{Results}
\label{sec:results}
A total of 188,324 human moves were compared with 155,394 GPT-2 agent moves. The quantities of moves by piece and the relative percentage of these moves can be seen in Table~\ref{tab:descriptive}.

\begin{table}[h]
	\centering
	\begin{tabular}{lllll}
		& \multicolumn{2}{c}{Counts} & \multicolumn{2}{c}{Percent} \\
		& Human        & GPT-2       & Human        & GPT-2        \\
		\midrule
		Pawns   & 49,386       & 51,408	      & 26.2\%       & 32.9\%       \\
		Rooks   & 31,507       & 25,997	      & 16.7\%       & 16.6\%       \\
		Bishops & 28,263       & 19,310	      & 15.0\%       & 12.4\%       \\
		Knights & 31,493       & 23,369	      & 16.7\%       & 14.9\%       \\
		Queen   & 22,818       & 16,260	      & 12.1\%       & 10.4\%       \\
		King    & 23,608       & 19,972	      & 12.5\%       & 12.8\%       \\
		\midrule
		Totals  & 188,324      & 156,316     & 100.0\%      & 100.0\%     \\
		\bottomrule
	\end{tabular}
	\caption{Descriptive statistics}
	\label{tab:descriptive}
\end{table}
The correlation between these populations is quite strong. A Pearson's two-tailed correlation coefficient is (97.794\%, 0.072\%). This strongly suggests that the GPT-2 model has internalized the movement biases of the players that it has been trained on. Further, Pearson's Chi-square test rejects the null hypothesis that the two populations are random with respect to each other with $p < 0.0001$.

\subsection{Move Legality}
\label{subsec:legality}
It is important to stress that the goal of this effort was not to train the GPT to play chess. Rather, the goal is to use chess as a ground truth mechanism to deduce the fidelity of the GPT in encoding \textit{human} beliefs. Critically, this involves attempting to understand when the model performs in a manner inconsistent with the known ground truth -- in this case, the board and the pieces. Although the model never moved a piece off the board (e.g. move a rook to a non-existent square \enquote{i9}), it did make incorrect moves occasionally.

A program was written to test the legality of all moves. For example, pawns can only move forward, their first move can be two squares, and they move diagonally when taking an opponent's piece. Each move in the database was tested against the rules for moves by piece. Because of the possibility that the model could learn from incorrectly transcribed games, the human games were also analyzed using the same technique. No errors were found in the transcribed human games. The results for the model are shown in Table \ref{tab:legal}.

\begin{table}[h]
	\centering
	\begin{tabular}{llll}
		& Illegal & Total   & Percent \\
		\midrule
		Pawns   & 14      & 51,408  & 0.03\%  \\
		Rooks   & 35      & 25,997  & 0.13\%  \\
		Bishops & 68      & 19,310  & 0.35\%  \\
		Knights & 35      & 23,369  & 0.15\%  \\
		Queen   & 332     & 16,260  & 2.04\%  \\
		King    & 19      & 19,972  & 0.10\%  \\
		\midrule
		Totals  & 503     & 156,316 & 0.32\% \\
		\bottomrule
	\end{tabular}
	\caption{Illegal GPT-2 moves}
	\label{tab:legal}
\end{table}

The number of errors that the GPT commits are low, but also seem to be proportional to the number of training moves, and the degrees of freedom (DOF) for each piece. The lowest percentage error is for the pawns, which are approximately 30\% of all human moves, and can only move to a small set of nearby squares, based on the color of the piece for a total of 8 DOF. The queen, on the other hand, is much more mobile, capable of moving to any one of 24 squares at one time, or 24 DOF. Other pieces fit onto the spectrum. The king, with 8 DOF but 12\% of the total moves, is closer to the pawns. An interesting case is that of the bishops and rooks, which have the same DOFs (16) and similar move percentages. However, bishops have a considerably higher error rate. It seems reasonable to assume that this a product of the row and column changing during a move for a bishop, rather than just the row or column changing for the rook. This would make the bishop behavior a more complex pattern for the model to infer. 

The results of the move legality of the ablation models trained on the smaller corpora are shown in Table~\ref{tab:legal_ablation}. As might be expected, movement errors occur more often with models trained on smaller corpora.  

\begin{table}[h]
	\centering
	\begin{tabular}{llllll}
		& 800k & 400k  & 200k & 100k & 50k \\
		\midrule
		Pawns   & 0.03\%  & 0.08\% & 0.19\% & 0.34\% & 0.26\% \\
		Rooks   & 0.13\%  & 0.07\% & 0.04\% & 0.08\% & 0.06\% \\
		Bishops & 0.35\%  & 0.47\% & 8.86\% & 34.91\% & 63.13\% \\
		Knights & 0.15\%  & 0.16\% & 2.30\% & 17.10\% & 29.01\% \\
		Queen   & 2.04\%  & 5.94\% & 20.29\% & 29.57\% & 38.13\% \\
		King    & 0.10\%  & 0.05\% & 0.11\% & 0.24\% & 1.38\% \\
		\midrule
		Average  & 0.47\% & 1.13\% & 5.30\% & 13.71\% & 22.00\% \\
		\bottomrule
	\end{tabular}
	\caption{Ablation models illegal moves}
	\label{tab:legal_ablation}
\end{table}

Significantly, all models were able to maintain percentages of moves by piece that were strongly correlated with the recorded human behavior from \textit{The Week in Chess}. Though the moves were more likely to be wrong, the percentage of overall moves appears to be resilient with respect to model size. 

Based on these results, it seems clear that the GPT is capable of encoding complex belief structures, such as the board, rules, and play structure of games such as chess. Given that this is true, what can be done with this knowledge?

\subsection{Navigation}
Relationships to an environment, mediated through simple rules is described in the parable of Simon's Ant. In it, Herbert Simon stated that the complex path traced by an agent (an ant in this case) as it exhibited complex behavior did not require a complex algorithm. Rather, the ant would apply a set of simple rules to the world as it experiences it at the moment. If there is an obstacle, it will attempt to go around it. If there is a threat, it will flee. If there is food, it will grab it an bring it back to the nest. These rules, given a particular environment, will produce an appropriate path~\cite{simon1996sciences}. This rule can be states as follows:

\begin{equation} \label{eq:simons_ant}
	Behavior \approx f(Rules | Environment)
\end{equation}

The agents that are created as a series of probes and their subsequent trajectory through the language space of the model also exhibit complex behavior, again based on simple rules such as the selected text and the length of the textual probe. This suggests that it is possible to infer the environment, it we rearrange the terms:

\begin{equation} \label{eq:simons_environment}
	Environment \approx f(Behavior | Rules)
\end{equation}

Once relationships between elements of the environment can be described, it is possible to \textit{project} them onto lower-dimensional spaces or \textit{maps}.

One of the main uses of maps is to support deliberate navigation -- to determine your current location, your desired destination, and to plot a course that makes the destination reachable using the means available. For automobiles, this would be a roadmap, often digital and updated automatically. For hiking, a topographic map showing trails can be essential. In the case of the chessboard, the mechanism of movement is the piece and the rules that govern its behavior. The question of navigation across a physical environment can be different from navigating a network. In the case of a network, distant nodes may still be linked. For example, the diagonal paths shown in Figure~\ref{fig:preserved_orthogonality}  would not be the shortest possible network path. That would be a two move sequence (assuming no blocking pieces), either by the rook (eg. a1, a8, h8) or the queen (e.g. d1, a1, h8). The shortest \textit{physical} distance is a move sequence that is closest to the diagonal that connects a1 to h8. This ability to explore belief spaces (in this case, the model's understanding of pieces and the board) to \textit{afford} a physical style of navigation is a primary goal of this research.

\begin{figure}[h]
	\centering
	\fbox{\includegraphics[width=0.9\linewidth]{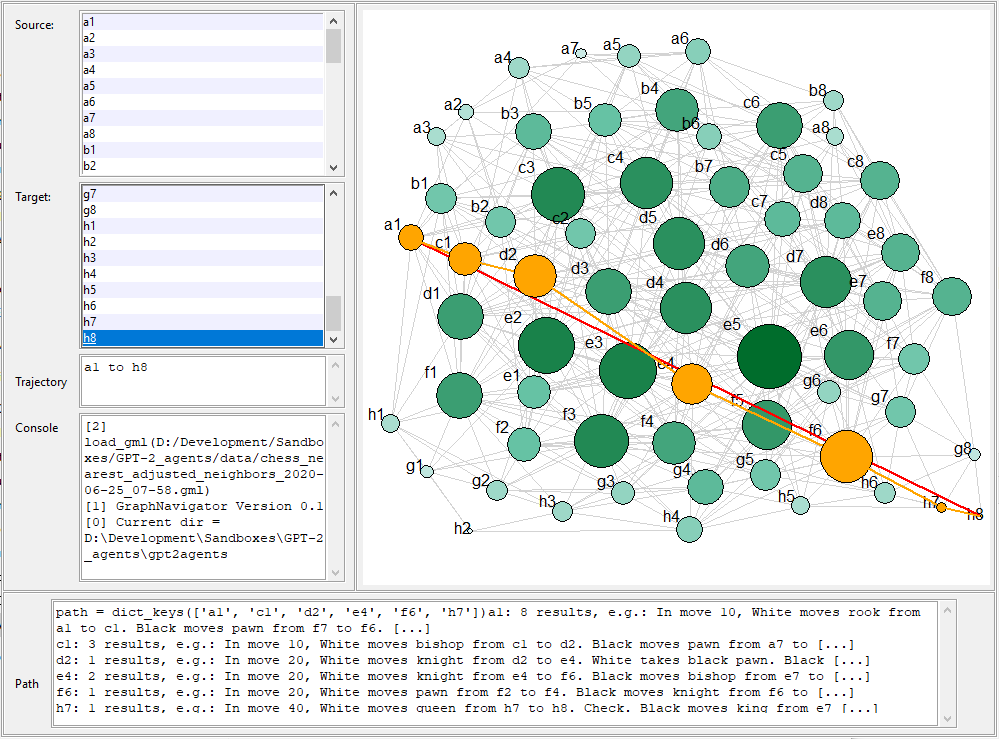}}
	\caption[]{Tool showing coarse navigation with long jumps}
	\label{fig:coarse}
\end{figure}

\begin{figure}[h]
	\centering
	\fbox{\includegraphics[width=0.9\linewidth]{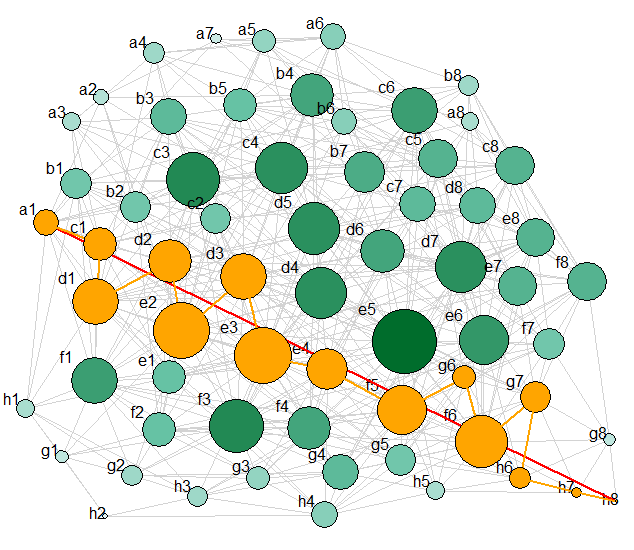}}
	\caption[]{Granular navigation with short jumps}
	\label{fig:granular}
\end{figure}

To do this, an interactive tool (Figure~\ref{fig:coarse}) was built that uses the physical locations of the nodes and their network connectivity together. The user can select any two nodes in the network, at which point a line is drawn that connects the origin and destination nodes in physical space. The user can then select a \textit{coarse} (Few jumps -- Figure~\ref{fig:coarse}) or \textit{granular} (Short distance -- Figure~\ref{fig:granular}) course to be plotted. The algorithm for finding the route is shown in Algorithm~\ref{alg:coarse}.

\begin{algorithm}[h]
	Set starting node $N_{cur}$\\
	Set target node $N_{tgt}$\\
	Create empty list of path nodes $L_{path}$\\
	\While{Ncur != Ntgt}{
		Set trajectory $T$ from $N_{cur}$ to $N_{tgt}$\\
		Distance $D_{prev}$ = $max$\\
		\ForEach{Node $N_{edge}$ connected to $N_{cur}$}{
			Compute closest point $P$ on $T$ to $N_{edge}$\\
			$D_{new}$ = distance from $N_{edge}$ to $P$\\
			\If{(P between $N_{cur}$ and $N_{tgt}$) and ($D_{new} < D_{prev}$)}{
				$N_{cur} = N_{edge}$\\
				$D_{prev} = D_{new}$\\
				Add $N_{cur}$ to $L_{path}$\\
			}
		}
	}
	\Return $L_{path}$\\
	\vspace{0.5em}
	\caption{Coarse path generation}
	\label{alg:coarse}
\end{algorithm}

This approach creates a path of nodes that are closest to the desired trajectory. However, there may be long jumps that may be more conceptually difficult for someone navigating a belief space. In this chess belief space, these longer jumps are accomplished by using the knight, which has a more complicated movement pattern than any of the other pieces, which move linearly. One might generalize that long jumps (what we might anecdotally refer to as conceptual leaps) may be more difficult to comprehend than smaller, incremental steps. To provide this alternative, The \textit{coarse} algorithm is modified to compute the distance as the square root of the sum of the distance to the node $Nedge$ and the trajectory $T$ (Equation~\ref{eq:granular_dist}, where $D_{Nedge}$ is the distance from $N_{cur}$ to $N_{edge}$, and $P$ is the closest point on the trajectory line.).

\begin{equation}
	\label{eq:granular_dist}
	D_{new} = \sqrt{D_{Nedge}^2 + D_P^2}
\end{equation}

This results in the generation of the \textit{granular} path shown in Figure~\ref{fig:granular}, where the nodes may be further from the line $T$, but are connected by shorter jumps, typically of only one square.

\begin{figure}[h]
	\centering
	\fbox{\includegraphics[width=.9\linewidth]{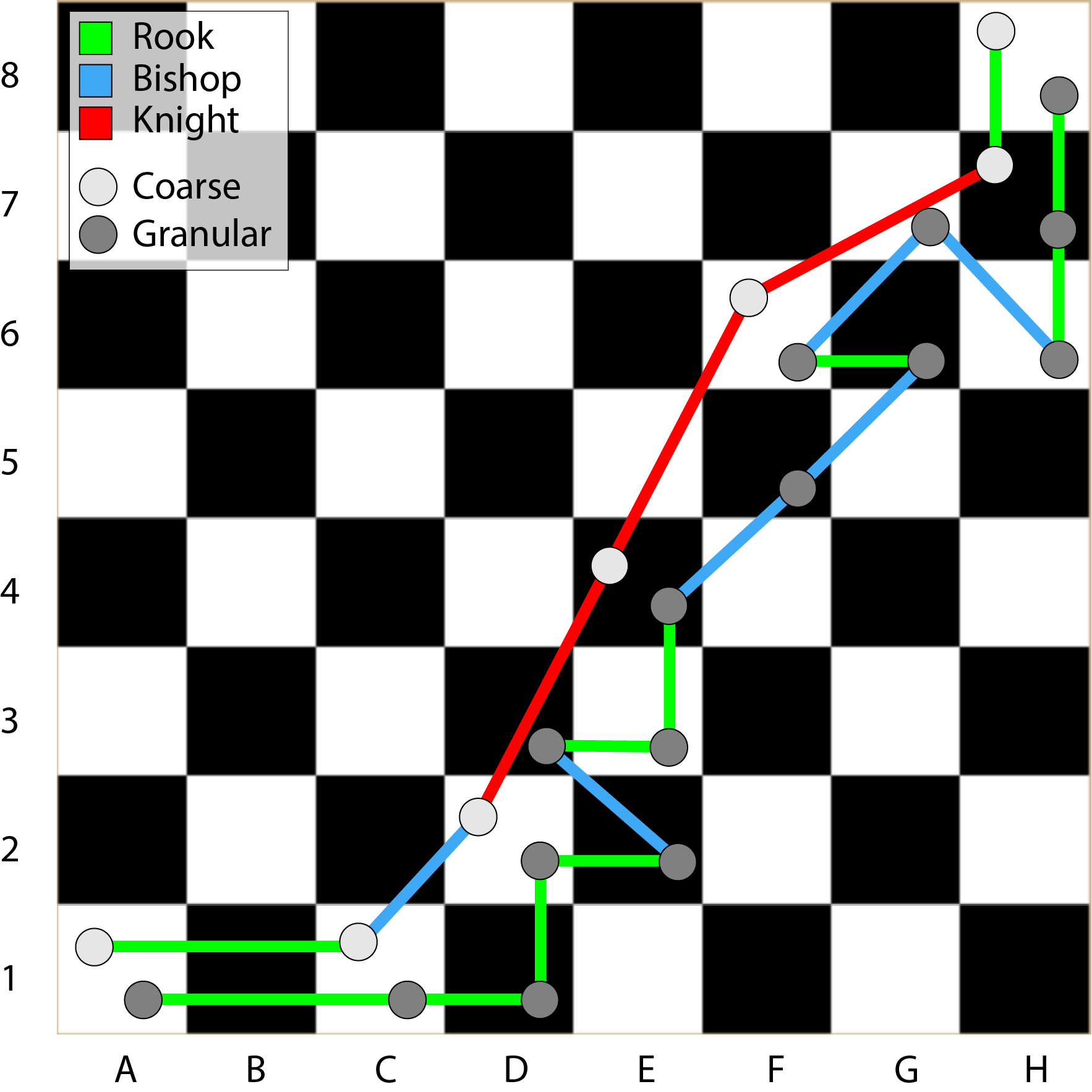}}
	\caption[]{Coarse and granular a1-h8 paths on chessboard}
	\label{fig:board}
\end{figure}

To see how these trajectories map from the constructed map to the ground truth of the chessboard, each set of moves was plotted in Figure~\ref{fig:board}. The color of the dots indicate the \textit{coarse} or \textit{granular} path, while the color of the line indicate the piece capable of the move. It should be noted that in addition to the rook and bishop, the queen and king can move horizontally and diagonally, while the pawn can move vertically and diagonally, when taking a piece. In this figure, the use of the knight for long jumps described earlier (Figure~\ref{alg:coarse}) can be seen clearly. 

\begin{figure}[h]
	\centering
	\fbox{\includegraphics[width=0.9\linewidth]{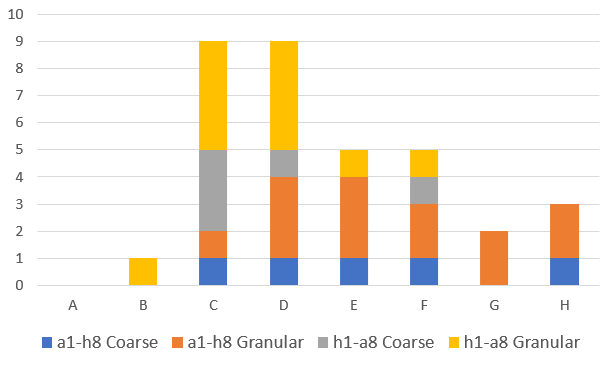}}
	\caption[]{Column occupation counts}
	\label{fig:columns}
\end{figure}

A bias in white player behavior can also be seen in in the way the moves tend to cluster in the center columns of the board (Figure~\ref{fig:columns}) in both the coarse and granular paths shown in Figure~\ref{fig:board} and in additional coarse and granular paths computed between a8-h1 (The a8-h1 trajectory is shown in Figure~\ref{fig:preserved_orthogonality}).  In all these paths, the first move is a horizontal move of two squares by the rook from the corner to the c1/f1 square. This reflects the behavior of the human players as captured by the language model and traversed by the agents. As such,it is an example of how trajectories compatible with human belief as encoded in actions, memorized by language models can be used to determine useful trajectories across belief spaces.

%% file: discussion.tex
\section{Discussion}
Language models, such as the GPT-2/GPT-3 encode and preserve human-generated information in a way that embeds the associations in fixed relationships. These models offer the opportunity to understand human belief and biases in novel ways. While the general models, trained on millions of web pages are impossible to validate with respect to known ground truth, models adapted through finetuning can be. In this study, we focused on validation using chess, but the results should be more broadly applicable.

\begin{figure}[h]
	\centering
	\begin{minipage}{.5\linewidth}
		\centering
		\includegraphics[width=.95\linewidth]{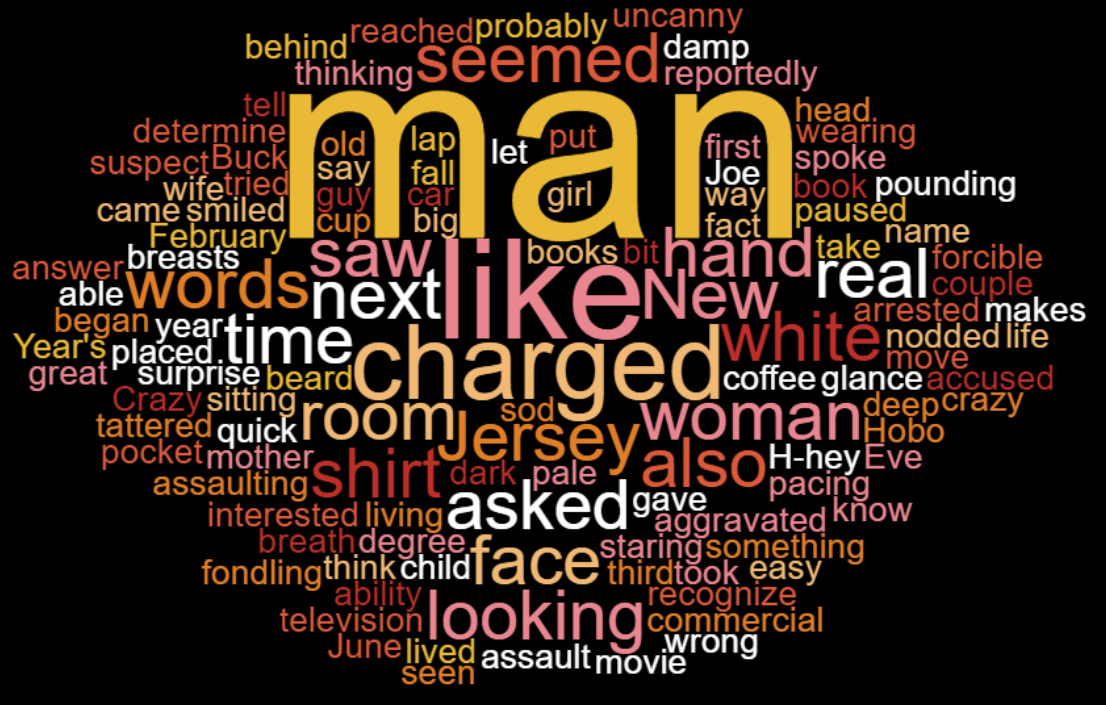}
	\end{minipage}%
	\begin{minipage}{.5\linewidth}
		\centering
		\includegraphics[width=.95\linewidth]{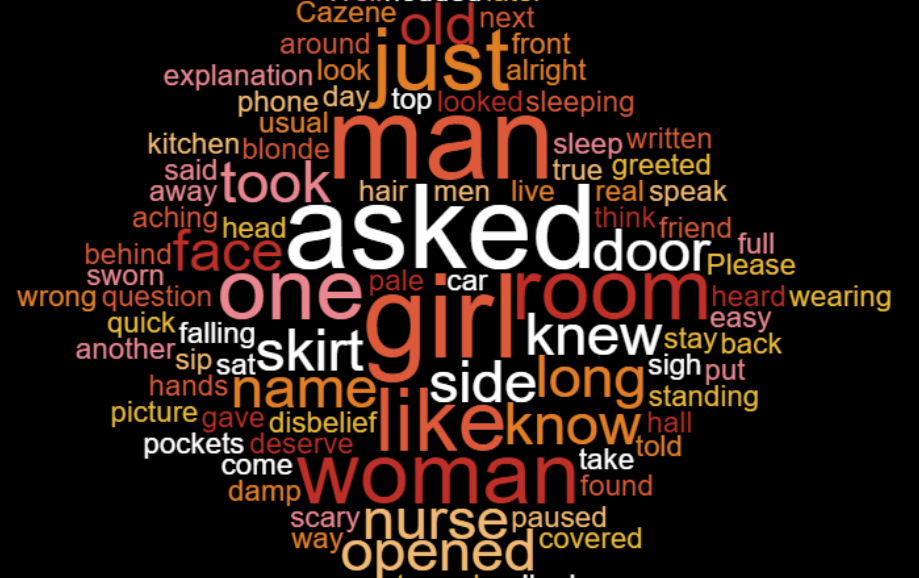}
	\end{minipage}
	\caption{\enquote{The man/woman walked into the room} word clouds}
	\label{fig:word_clouds}
\end{figure}

In these models, textual agents, using the same prompt with a different random seed will produce a population of responses that define the region around the query. In many respects the use of such agents to explore these spaces allow whole populations to be rigorously sampled in ways that surveys and text mining have not been able to support. For example, let's revisit the text generation exercise from the introduction. Because the same prompt (\enquote{\textbf{The man walked into the room. }}) can be run thousands of times, statistical patterns emerge that can be captured in a variety of visualizations, such as word clouds in Figure \ref{fig:word_clouds}. For these visualizations, the two responses from before and after the texts were included. Some patterns become evident even on casual inspections, such as the emphasis on \enquote{man} in the cloud on the left, with no corresponding \enquote{woman} on the right. 

More sophisticated displays such as maps also become accessible with this technique. Maps portray a continuous and unique spatial relationship between the elements within the borders of the map. This differs word clouds, network graphs, timelines, and charts because the continuous spatial relationship means that a line drawn in any orientation on the map represents a meaningful trajectory in the space that the map portrays. 

The statistical validation of the model's generated description of chess moves against human gameplay has two implications for the design of studies using this technique. Data that represents large scale structures in human belief such as the overall percentage of move by piece in chess appear to be represented in models trained on less data. This implies that broad questions may be realistically be answers with comparatively small corpora. However, answers that require a high degree of detail benefit dramatically from more data. 

Beyond the statistical validation of the models, we were also able to produce two-dimensional maps of the chessboard and plot trajectories across it. The paths generated to follow these plots connected together moves in novel, and overwhelmingly \textit{legal} ways that supported navigation in the space as encoded in the language model. It is reasonable to expect that these same techniques -- corpus extraction, finetuning, probing the model using repeated textual prompts, and path creation should apply to other, less structured human data. 

This implies a new set of mechanisms for fields that study the behavior of populations, such as sociology. Synthetic agents can be \enquote{launched} into language models to determine with repeatable precision, the state and relationships of human belief in narrow (e.g. games) and potentially broad (e.g. Twitter, Reddit, etc) domains. 

An important point to discuss here is the role that traditional qualitative research plays in an environment that includes language models such as those discussed in this paper. It is critical to understand that most if not all deep learning involves a generalization process. Objective functions tend to emphasize broader relationships, and outliers that exist in the world may be too sparse to be picked up in the model(s). Machine learning applications like this should be able to provide broad outlines down to some level of granularity, but for smaller, more intimate patterns, there is unlikely to ever be a substitute for the human researcher. 

%% file: future_work.tex
\section{Future Work}
The idea of using probes into a language model as \enquote{agents} in a human-defined belief space should be broadly applicable. Once a large enough text from the population in question has been gathered, a model can be trained. These trained models support flexible interactions. Information needs that emerge from early interactions with the model can be met through the use of new textual probes. Results are consistent and repeatable, and model accuracy at different levels of granularity increase with proportion to the size of the corpora. 

Using such a mechanism allows for an agent/human hybrid form of sociology. Data gathered from humans is used to produce a fitness landscape of textual relationships that neural NLP agents navigate. Since the agents are not human, data is effectively generalized and anonymized. That being said, document similarity measures could be used to find, for example, tweets from human users that have a high similarity to the synthetic tweets generated by a model trained on a Twitter corpora.

We are currently looking at using this approach on two Twitter datasets that started in December 2019, one involving COVID-19 related tags and one storing racism-related tags. Training models on these datasets should allow for an initial mapping of these textual spaces. One of the patterns that we are particularly interested in mapping is how the concept of wearing masks in the USA went from fringe, to accepted by most, to polarizing. We believe that by mapping this path across the belief space of the topic may provide insight into associated beliefs as well as insight into where this trend in behavior may be headed.

Large social events that generate millions of lines of text do not appear to be needed for this approach. Given the accuracy in legal moves for the 800k line chess model, smaller scale experiments should be achievable. For example, one of our projects involves the exploration of urban mobility approaches in Trento, Italy. Using a hybrid approach of programmatic text generation (e.g. \enquote{Ms. Abc requested a car to travel from Xxx to Yyy} and user data from social media posts and interactions between transportation providers and consumers. A model could be built to discover and understand the subtle relationships between people's transportation-related needs and the services that they use. We are particularly intrigued by the use of hybrid systems to generate text that can richness and detail to the sort of actions that human users tend not to document.

We are also looking at more sophisticated mechanisms for extracting topics and building the network from which the graphs are made. Because data from social networks can be very large, scaling would be an issue with our current techniques. One of the most promising areas that we wish to explore are the creation of knowledge graphs using neural embedding techniques \cite{yang2014embedding, bosselut2019dynamic}, and embeddings combined with tensor factorization \cite{kazemi2018simple}.  These approaches manage the growth in complexity with respect to the size of the embeddings. 

Lastly, user tools and visualizations are important parts of this research. Creating navigable, interactive 2D and 3D displays will continue to be developed, along with mechanisms that tie the text of the statements made by the model into the user interface.

%% file: conclusions.tex
\section{Conclusions}

In this article, we have described a framework that supports the creation of maps of belief space that can be shown to be correct when compared against the ground truth encoded in the structure of a GPT-2 Language model fine-tuned on chess. 

The 117M-parameter GPT-2 model was trained on text describing 23,000 chess games hosted on theweekinchess.com. A variety of textual prompt-based agents were repeatedly run against the model, and the results were parsed and stored in a database. A statistical analysis was performed comparing the spectral characteristics of piece movement of historic human and synthesized chess game description. These populations were found to be statistically similar with a $> 97\%$ probability.  

Using the agent-generated chess piece moves relationships, a network was created and laid out using a directed force approach. This layout largely preserved the ground truth of the original chessboard, and could be used to to support the plotting of trajectories between the two points and the creation of paths that consisted of legal moves that connected the endpoints of the trajectory. These paths reflected the bias of the original human data, such as concentrating moves in the center columns of the board.

Lastly, further applications beyond such constrained domains were discussed, with particular emphasis on using social media corpora (including programmatically enhanced social media) to explore and create maps of less structured domains.

